\pdfoutput=1

\documentclass[11pt]{article}

\usepackage[]{acl}

\usepackage{times}
\usepackage{latexsym}
\usepackage{booktabs}
\usepackage{multicol}
\usepackage{comment}
\usepackage{covington}
\usepackage[T1]{fontenc}

\usepackage[utf8]{inputenc}

\usepackage{microtype}

\title{Annotating Norwegian Language Varieties on Twitter for Part-of-Speech}

\author{Petter Mæhlum$^1$, Andre Kåsen$^2$, Samia Touileb$^3$ and Jeremy Barnes$^4$ \\
\\
$^1$University of Oslo\\
$^2$National Library of Norway\\
$^3$University of Bergen\\
$^4$University of the Basque Country\\
  \\
   \texttt{pettemae@ifi.uio.no}, \texttt{andre.kasen@nb.no},\\ \texttt{samia.touileb@uib.no}, \texttt{jeremy.barnes@ehu.eus}
  \\
  }

\begin{document}
\maketitle
\begin{abstract}
Norwegian Twitter data poses an interesting challenge for Natural Language Processing (NLP) tasks. These texts are difficult for models trained on standardized text in one of the two Norwegian written forms (Bokmål and Nynorsk), as they contain both the typical variation of social media text, as well as a large amount of dialectal variety. In this paper we present a novel Norwegian Twitter dataset annotated with POS-tags. We show that models trained on Universal Dependency (UD) data perform worse when evaluated against this dataset, and that models trained on Bokmål generally perform better than those trained on Nynorsk. We also see that performance on dialectal tweets is comparable to the written standards for some models. Finally we perform a detailed analysis of the errors that models commonly make on this data.
\end{abstract}

\section{Introduction}
Norwegian Twitter data poses an interesting challenge for Natural Language Processing (NLP) tasks. Not only do these data represent a set of noisy, user-generated texts with the kinds of orthographic variation common on social media, but also because there is a considerable number of tweets written in dialectal Norwegian. These dialectal variants are quite common and add another level of difficulty for NLP models trained on clean data in one of the two Norwegian written forms (Bokmål or Nynorsk).

\newcite{barnes-etal-2021-nordial} compiled a dataset of tweets classified according to whether they are written in primarily Bokmål, Nynorsk, or a dialect of Norwegian. We build upon this work by annotating a subset for Part-of-Speech (POS). We investigate to what extent available Norwegian POS tagging models, that were trained on Bokmål and Nynorsk Universal Dependency data \cite{nivre-etal-2020-universal}, perform on this Twitter dataset.

To this end, we use five POS models: three off-the-shelf models, and two developed for the purpose of this work. Each of these models was trained on either a dataset of Bokmål or Nynorsk texts. We explore the performance of each model in terms of accuracy, and investigate which standardized written form can be used as training data and yield good results for non-standardized dialectal texts. 

The main contributions of this work are:

\begin{itemize}\itemsep0.2em
    \item we annotate a moderately sized Twitter dataset with POS labels and include metadata related to which language variety it belongs (Bokmål, Nynorsk, Dialect, or Mixed),
    \item we perform a detailed error analysis of common model errors specific to our Twitter data,
    \item we include our insights into the annotation process for POS tagging of non-standardized written forms,
    \item we release two spaCy models built on top of a Norwegian BERT model.
\end{itemize}

\section{Background}\label{sec:background}
\citet{Joh:90} outlined a system for automatic morphosyntactic analysis of Norwegian nouns in the framework of \citet{Kos:83}. This was among the first systems, if not even the very first, that automatically assigned Norwegian texts any morphological information. The first widely used tagger, however, was developed within the \textit{Taggerprosjektet}\footnote{The project ran from April 1996 to December 1998.} and came to be known as the Oslo-Bergen Tagger\footnote{\url{https://github.com/noklesta/The-Oslo-Bergen-Tagger}} (OBT). Rather than continuing and expanding the system of \citet{Joh:90}, OBT was implemented in the framework of \citet{Kar:90}. OBT was initially a rule-based Constraint Grammar tagger for Norwegian Bokmål. Later, both support for Norwegian Nynorsk and a statistical disambiguation component were added \cite{Joh:Hag:Lyn:12}. But one drawback of OBT is that it is made for written, edited text, and therefore might not scale well to sources that are not standardised.

Extending tagger coverage to spoken Norwegian dialect transcription, on the other hand, was the objective of both \citet{Nok:Sof:07} and \citet{Kas:Hag:Nok:Pri:19}. Both sampled data either from the Norwegian part of the Nordic Dialect Corpus (NDC, \citet{NDC:09}) or the Language Infrastructure made Accessible (LIA) Corpus.\footnote{\url{https://tekstlab.uio.no/LIA/korpus.html}} Annotations are found in the respective treebanks of the corpora and are accounted for in \citet{Oev:Kas:Hag:Nok:Sol:Joh:18} and \citet{Kas:Hag:Nok:Pri:Sol:Hau:22}. 

Besides Norwegian, there is a large amount of work on the difficulty of processing noisy data from social media \cite{ws-2015-noisy}, including the difficulty of POS tagging on social media \cite{albogamy-ramasy-2015-towards}, with dialectal variation \cite{jorgensen-etal-2015-challenges}, or whether lexical normalization is helpful \cite{van-der-goot-etal-2017-normalize}. However, Norwegian currently lacks any of these studies.

\section{Data}\label{sec:data}
Resources for evaluating NLP pipeline tasks for Norwegian are scarce. The only dataset available for standard NLP tasks such as POS tagging, lemmatization, and parsing is the Norwegian Dependency Treebank (NDT, \citet{Sol:13}, \citet{Sol:Skj:Oev:Hag:Joh:14}) that has been converted to the Universal Dependencies standard \citep{Oev:Hoh:16}. There is, however, a notable exception when it comes to transcribed spoken dialectal data, where the LIA and NDC treebanks as mentioned above are available with annotations for POS tags, morphological features, lemmas, and dependency-style syntax. Despite this, the transcribed texts in the LIA and NDC corpora do not share the same characteristics as the Twitter data. Twitter contains spelling errors and emoji,\footnote{Emoji has recently gained some interest in the linguistic literature (see \url{https://ling.auf.net/lingbuzz/005981})} along with mentions and hashtags. We observe that although our Twitter data contains some characteristics of spoken Norwegian, such as subjectless sentences as in \ref{ex:subjectless}, which is otherwise within the spelling norms, the spelling conventions differ from those of LIA and NDC, making it difficult to directly compare the data.

\begin{covexample}
\gll  Kommer nok hjem snart .
Comes probably home soon .
\glt‘(Unspecified) probably comes home soon .’
\glend
\label{ex:subjectless}
\end{covexample}

In LIA and NDC, all transcriptions are done according to a Norwegian-based semi-phonetic standard \cite{Hag:Haa:Ols:Sof:15}, with strict marking of vowel quantity, palatalization, retroflexion, and more. We see that writers on Twitter do not conform to any specific spelling norm when writing in their own or another dialect. This means that although not all dialectal traits from a dialect are faithfully preserved, this still leads to much dialectal variation in the Twitter data, as things that could have had a common spelling is spelled according to the author's own preference. Especially phonetic differences are often not indicated on Twitter.
Because of this, we needed a separate dataset that could be used to evaluate how various systems for Norwegian POS-tagging work on dialectal text as it is found on real data from social media platforms. 

We sampled a balanced subset of the dataset introduced by \citet{barnes-etal-2021-nordial}, who developed to develop a dialect classifier for Norwegian tweets, with the aim to be able to further investigate issues related to dealing with dialectal data on Twitter. This subset includes a selection of 38 tweets in Bokmål, 31 tweets in Nynorsk, and 35 in dialects, which comprises their full test set. We acknowledge that the size of the dataset is small. The POS-tagged dataset is subject to restrictions due to it containing personal information, but is available upon request.

\begin{table*}[t]
    \centering
    \begin{tabular}{lrrrrrrrrrr}
    \toprule
    & \multicolumn{2}{c}{\textit{Bokmål}} & \multicolumn{2}{c}{\textit{Nynorsk}} & \multicolumn{2}{c}{\textit{Dialectal}} & \multicolumn{2}{c}{\textit{Mixed}} & \multicolumn{2}{c}{\textit{All}} \\
    \cmidrule(lr){0-0}\cmidrule(lr){2-3}\cmidrule(lr){4-5}
    \cmidrule(lr){6-7}\cmidrule(lr){8-9}\cmidrule(lr){10-11}
PUNCT & 211 & 15.72\% & 151 & 13.92\% & 123 & 11.27\% & 35 & 13.46\% & 520 & 13.76\% \\
NOUN & 168 & 12.52\% & 150 & 13.82\% & 116 & 10.63\% & 37 & 14.23\% & 471 & 12.47\% \\
VERB & 157 & 11.7\% & 130 & 11.98\% & 129 & 11.82\% & 24 & 9.23\% & 440 & 11.65\% \\
PRON & 134 & 9.99\% & 97 & 8.94\% & 140 & 12.83\% & 29 & 11.15\% & 400 & 10.59\% \\
ADP & 120 & 8.94\% & 89 & 8.2\% & 85 & 7.79\% & 21 & 8.08\% & 315 & 8.34\% \\
AUX & 78 & 5.81\% & 84 & 7.74\% & 93 & 8.52\% & 19 & 7.31\% & 274 & 7.25\% \\
ADJ & 103 & 7.68\% & 67 & 6.18\% & 88 & 8.07\% & 13 & 5.0\% & 271 & 7.17\% \\
PROPN & 92 & 6.86\% & 74 & 6.82\% & 50 & 4.58\% & 18 & 6.92\% & 234 & 6.19\% \\
ADV & 77 & 5.74\% & 68 & 6.27\% & 74 & 6.78\% & 11 & 4.23\% & 230 & 6.09\% \\
SCONJ & 46 & 3.43\% & 42 & 3.87\% & 43 & 3.94\% & 5 & 1.92\% & 136 & 3.6\% \\
DET & 49 & 3.65\% & 38 & 3.5\% & 33 & 3.02\% & 14 & 5.38\% & 134 & 3.55\% \\
CCONJ & 34 & 2.53\% & 38 & 3.5\% & 44 & 4.03\% & 11 & 4.23\% & 127 & 3.36\% \\
PART & 36 & 2.68\% & 22 & 2.03\% & 29 & 2.66\% & 9 & 3.46\% & 96 & 2.54\% \\
X & 16 & 1.19\% & 15 & 1.38\% & 9 & 0.82\% & 10 & 3.85\% & 50 & 1.32\% \\
NUM & 11 & 0.82\% & 8 & 0.74\% & 14 & 1.28\% & 2 & 0.77\% & 35 & 0.93\% \\
INTJ & 7 & 0.52\% & 6 & 0.55\% & 14 & 1.28\% & 1 & 0.38\% & 28 & 0.74\% \\
SYM & 3 & 0.22\% & 6 & 0.55\% & 7 & 0.64\% & 1 & 0.38\% & 17 & 0.45\% \\
    \bottomrule
    \end{tabular}
    \caption{Distribution of each POS-tag in the Twitter test set, along with the total number of occurrences for each tag and their corresponding percent-wise distribution.}
    \label{tab:postags}
\end{table*}

\subsection{Norwegian Dialects}
Norwegian is considered to have four main dialect groups based on four different traits. This has been a controversial matter and the four-way divide essentially follows \citet{Chr:54}. There are also recent proponents of a two-way divide \cite{Skj:97}. The four-way distinctions have a Northern, Middle, Western, and Eastern group, whereas the two-way divide only operates with a Western and Eastern group. But these distinctions are made with traits from the spoken language. And, as \citet[p. 29]{Mae:Roy:12} point out, there is a discrepancy between how dialectologists and lay people classify dialects. What sort of dialectal traits Twitter users choose to include may therefore lead to a different kind of divide than one can find in the dialectology literature. That being said, \citet{Ven:90} shows that there has been a long tradition of writing in dialect, where the oldest text in \citet{Ven:90} dates back to 1525.

\subsection{POS Annotations}
The texts from the test set were annotated using the Universal Dependencies POS tagset.\footnote{https://universaldependencies.org/u/pos/} The tweets were tokenized with NLTK's tokenizer  \citep{bird2009natural} and split into sentences manually. The NLTK tokenizer was chosen over other tokenizers as our preliminary testing on our Twitter dataset shows that it performs better on noisy Norwegian data. The tokenized data was then pre-annotated with Stanza's Bokmål tokenizer to alleviate the annotation task. The remaining task was to correct each POS-tag for these pre-annotated sentences. One annotator annotated the whole test set, while two other annotators annotated two separate subsets of the dataset to give an indication of how robust the annotations were. All three annotators were trained in linguistics and language technology, and are native Norwegian speakers. An overview of the distribution of each POS-tag for each written form is reported in table \ref{tab:postags}. We see that the percent-wise distribution of POS-tags is similar in Bokmål, Nynorsk and All, but that the PRON tag is somewhat more frequent than the VERB tag in the Dialect tweets. This could be due to the fact that some dialectal tweets only appear as dialectal due to specific dialectal pronouns.

\subsection{Inter-Annotator Agreement}
The inter-annotator score for the full doubly-annotated test set, using Cohen's $\kappa$, was 0.87, indicating quite high agreement. Looking at the specific categories, we see that the agreement was 0.92 for Bokmål, 0.83 for Nynorsk, and 0.88 for dialectal tweets. No specific error patterns are observed that would account for the difference in scores, but all annotators have more familiarity with the Bokmål variant. One common point of disagreement across all is the copula verb \textit{å være} `to be', which according to the UD guidelines should be tagged as AUX. This was commonly tagged as VERB by one of the annotators. There is also some disagreement when it comes to words such as \textit{opp} `up', and \textit{ned} `down', which can be tagged both as adverbs (ADV), adpositions (ADP), and verbal particles. Since there is no tag for verbal particles in UD, the annotators had to chose between the other two. Cases of disagreement were solved by discussing tags where one or more annotators disagreed.

\section{Experiments}\label{sec:experiments}

We test several models trained on available Norwegian UD datasets on our Twitter data. Specifically, we compare OBT, Stanza, UDPipe 2.0, a simple BiLSTM model, as well as training our own spaCy models. 

Both Stanza \citep{qi-etal-2020-stanza} and UDPipe 2.0 \citep{straka-2018-udpipe} use a BiLSTM which takes features from 1) pre-trained word embeddings, 2) a trainable frequent word embedding that is randomly initialized before training, and 3) character-level LSTM features. While UDPipe only uses a softmax layer for classification, Stanza instead uses a biaffine classifier to ensure consistency between the UPOS and XPOS predictions.

The BiLSTM model we use is a simplified version of the models used in UDPipe and Stanza. The model does not take any pre-trained word embeddings as features, but rather uses the vocabulary of the dataset it is trained on to create the embeddings. The model uses a linear layer for classification.

The spaCy models are newly trained during the present work, and will be released publicly in the near future. Since spaCy is a fully configurable and trainable pipeline, we used the Norwegian BERT model described in \cite{Kum:Ros:Wet:Bry:21} with a shared embedding layer for a tagger, morphologizer, and trainable lemmatizer in an effort to optimize the tagger task.

\section{Results and Discussion}\label{sec:results}

\begin{table*}[ht]
    \centering
    \begin{tabular}{lrrrr}
    \toprule
    & Bokmål & Nynorsk & Dialect & All \\
    \cmidrule(lr){1-1}\cmidrule(lr){2-2}\cmidrule(lr){3-3}\cmidrule(lr){4-4}\cmidrule(lr){5-5}
        OBT Bokmål & 77.8  & - & 62.3 & -  \\
        OBT Nynorsk & -  & 73.1 & 57.3 & -  \\
        BiLSTM\_UD Bokmål & 80.5 & 63.8 & 63.4 & 70.2 \\
        BiLSTM\_UD Nynorsk & 62.3 & 76.2 & 56.7 & 64.6  \\
        BiLSTM\_UD Nynorsk\_LIA & 47.6 & 56.2 & 43.9 & 48.9 \\
        Stanza Bokmål & 86.6  & 67.5 & 69.5 & 75.4  \\
        Stanza Nynorsk & 45.8  & 82.8 & 52.0 & 58.1  \\
        UDPipe Bokmål & \textbf{89.6}  & 76.1 & 72.9 & 80.4  \\
        UDPipe Nynorsk & 74.4  & 82.9 & 63.2 & 73.2  \\
        spaCy Bokmål & 87.9  & \textbf{85.7} & \textbf{83.3} & \textbf{85.8}  \\
        spacy Nynorsk & 62.5  & 83.2 & 65.0 & 69.6  \\
    \bottomrule
    \end{tabular}
    \caption{Accuracy on our Twitter test set using five different models trained on either Bokmål or Nynorsk datasets.}
    \label{tab:results}
\end{table*}

Table \ref{tab:results} gives an overview of the accuracy on the Twitter test set using our five models trained on either Bokmål or Nynorsk data. Note that due to their small number, we do not include the mixed category by itself, but these tweets are included in the ALL column.
On our twitter Bokmål test set, the best model is the UDPipe Bokmål model, which achieves 89.6 accuracy. Generally, the models trained on the UD Bokmål data are consistently better than the Nynorsk versions on this data (an average of 26.5 percentage points (pp)). 
Interestingly, the same is not true for the Twitter Nynorsk data. One may assume that models trained on the Nynorsk UD data would always perform better, but in fact, the best performing model is the spaCy model trained on Bokmål (85.7 acc) and on average, the models trained on UD Bokmål perform 4.9 pp worse.

Finally, on the dialectal Twitter data, the spaCy Bokmål model once again performs best (83.3). Again training on the Bokmål data generally performs 12.8 pp better than training on Nynorsk data. This may be due to the subset of dialectal tweets, as a manual inspection showed a large number of tweets from Central and Northern dialects, which share more features with Bokmål. A larger number of tweets from Western and Southern dialects could potentially change this. At the same time, however, it seems clear that the spaCy Bokmål model performs quite well on all the Twitter test data (85.8 acc), so it may simply be a stronger model.

\subsection{Error Analysis}

We note that the models struggle with features that are typical of the noisy Twitter data containing several misspellings. One concrete example is \textit{å}, which in normative writing most likely refers to the identically spelled infinitive marker \textit{å} `to'. However, as dialectal writing is much more relaxed, alternative spellings create new homographs that need to be dealt with. We see that some cases of `å' refer to the conjunction \textit{og} `and', which in many dialects is homophonous with \textit{å}.
We also note that many of the errors come from erroneously tagging pronouns as other word classes, such as INTJ, PART, or NOUN. One reason why there are many errors of this type might simply be because these are frequent indicators of dialect. \citet{barnes-etal-2021-nordial} show that certain pronouns such as \textit{æ} and \textit{mæ} (both `I') are highly correlated with dialectal tweets. They are in some cases the only dialectal indicator in a tweet.
Finally, we observe that there are problems with annotating enclitic elements and words that should have been written separately, or conversely, with compound words that have been split. The two latter problems are not exclusive to dialects, but are common in informal writing. Enclitic elements, such as the enclitic negation ('kke, 'kje, 'che, etc.) and enclitic pronouns such as \textit{'n} `he, him' and \textit{'a} `she, her' are sometimes added after words, and sometimes without any punctuation, and there are no tokenizers that the authors are aware of that can correctly separate out these enclitic elements. For example, a spelling like \textit{ekkje}, `is not', which is the copula \textit{e} with the enclitic negation adverb \textit{kkje} `not' written as one word, has this issue. The same happens with other words that according to the norm should be written as two words, such as \textit{i dag} `today', being written as \textit{idag}. This leads to tokens with multiple possible POS-tags. In these cases the annotators would consider what would be the best functional fit. For example, the resulting adverbial phrase \textit{idag} can be annotated as an adverb, and verbs negated by enclitics are tagged as verbs. However, these are the annotators' judgements, and their proper treatment is not clear from the UD guidelines.

\section{Conclusion}\label{sec:conclusion}

In this paper, we have introduced the first dataset of Norwegian tweets annotated for Part-of-Speech, that also include the metadata for the language variety of each tweet (Bokmål, Nynorsk, Dialect, or Mixed). We tested several POS taggers trained on UD data and show that, for our Twitter data, it is generally better to train on the UD Bokmål data, even if testing on Nynorsk or Dialect. Our detailed error analysis showed that the models generally have problems with dialectal pronouns and unfamiliar compounds. Finally, we release the newly trained spaCy models, and make our annotated data available on request, in order to enable the reproduction of our results.

\section*{Acknowledgements}

We thank Alexandra Wittemann and Annika Willoch Olstad for their valuable contributions as annotators. 

Parts of this work were supported by industry partners and the Research Council of Norway with funding to \textit{MediaFutures: Research Centre for Responsible Media Technology and Innovation}, through the centers for Research-based Innovation scheme, project number 309339.

\bibliography{anthology,custom}

\end{document}